\documentclass[letterpaper, 10 pt, conference]{ieeeconf}  

\IEEEoverridecommandlockouts                              

\usepackage{graphicx} 
\usepackage{epsfig} 
\usepackage{mathptmx} 
\usepackage{times} 
\usepackage{amsmath} 
\usepackage{amssymb}  
\usepackage{hyperref}
\hypersetup{
	pdftitle={},
	pdfsubject={},
	pdfauthor={},
	pdfkeywords={}
}
\usepackage{mathtools}
\usepackage{siunitx}
\usepackage{setspace}
\usepackage{nicefrac}
\usepackage{flushend}
\usepackage{changepage}

\usepackage{tikz}
\usepackage{setspace}
\usetikzlibrary{decorations.text}
\usetikzlibrary{decorations.markings}
\usetikzlibrary{calc}
\usetikzlibrary{shapes}
\usetikzlibrary{positioning}
\usetikzlibrary{arrows}
\usetikzlibrary{shapes.arrows}
\usetikzlibrary{arrows.meta}
\usetikzlibrary{fit}
\usetikzlibrary{shapes.geometric}
\usetikzlibrary{backgrounds}

\newlength{\minimumsizeaicon}
\newlength{\minimumheightaicon}
\newlength{\minimumsizeinterconnection}
\newlength{\minimumheightinterconnection}
\newlength{\minimumsizegoal}

\definecolor{aicRed}{HTML}{e78284}   
\definecolor{aicGreen}{HTML}{a6d189} 
\definecolor{aicBlue}{HTML}{8caaee}  
\definecolor{aicOrange}{HTML}{e5c890}  
\definecolor{aicGray}{HTML}{dce0e8}

\tikzset{
	aicon/state/.style={
		rectangle,
		fill=aicBlue,
		rounded corners=4pt,
		minimum width=\minimumsizeaicon,
		minimum height=\minimumheightaicon,
		inner sep=0pt,
		text=white,
		align=center,
		font=\linespread{0.8}\selectfont\small
	},
	aicon/active/.style={
		rectangle,
		rounded corners=4pt,
		fill=aicGreen,
		minimum width=\minimumsizeinterconnection,
		minimum height=\minimumheightinterconnection,
		inner sep=0pt,
		text=white,
		align=center,
		font=\linespread{0.8}\selectfont\footnotesize
	},
	aicon/sensor/.style={
		rectangle,
		rounded corners=4pt,
		fill=aicOrange,
		minimum width=\minimumsizeaicon,
		minimum height=\minimumheightaicon,
		inner sep=0pt,
		text=white,
		align=center,
		font=\linespread{0.8}\selectfont\small
	},
	aicon/goal/.style={
		diamond,
		fill=aicRed,
		minimum size=\minimumsizegoal,
		inner sep=0pt,
		text=white,
		font=\linespread{0.8}\selectfont\small
	},
	aicon/link/.style={
		draw=aicGray,
		line width=2.5pt,
	},
	aicon/gradient/.style={
		draw=aicRed,
		line width=2.5pt,
		->,
		>=latex
	}
}

\newcommand{\imgcoord}[4]{%
	\coordinate (imgpt@b) at ($(#1.south west)!#2!(#1.south east)$);%
	\coordinate (imgpt@t) at ($(#1.north west)!#2!(#1.north east)$);%
	\coordinate (#4) at ($(imgpt@b)!#3!(imgpt@t)$);%
}

\usepackage{cleveref}
\usepackage[noadjust,compress,sort,nospace,move]{cite}

\Crefname{figure}{Fig.}{Figs.}
\Crefname{section}{Sec.}{Secs.}
\Crefname{equation}{Eq.}{Eqs.}

\DeclareMathOperator*{\argmax}{argmax}
\DeclareMathOperator*{\Tr}{Tr}

\title{\LARGE \bf Resolving Conflicts Where and When They Arise:\\ Reactive Composition of Multi-Goal Behavior}

\author{Vito Mengers$^{1,2}$ \qquad\ \qquad\ \qquad Oliver Brock$^{1,2,3}$
	\thanks{$^1$ Robotics and Biology Laboratory, Technische Universit\"at Berlin}
	\thanks{$^2$ Science of Intelligence (SCIoI), Cluster of Excellence, Berlin, Germany}
	\thanks{$^3$ Robotics Institute Germany}
	\thanks{We gratefully acknowledge funding by the Deutsche Forschungsgemeinschaft (DFG, German Research Foundation) under Germany's Excellence Strategy -- EXC 2002/2 ``Science of Intelligence'' -- project number 390523135. This work has been partially supported by the German Federal Ministry of Research, Technology and Space (BMFTR) under the Robotics Institute Germany~(RIG).}}%

\begin{document}
\bstctlcite{IEEEexample:BSTcontrol}

\maketitle

\begin{abstract}
Multi-goal robotic tasks are commonly delegated to planning, because reactive control is prone to local minima when objectives conflict. We show that many such failures stem from static goal representations, not task complexity: the designer defines the objectives, but where and when they must be traded off follows from the world's interaction structure which changes with the state. To read this structure, we extend Active InterCONnect (AICON), a graph of recursive estimators coupled by differentiable interconnections, with adaptive nullspace projections: lower-priority gradients enter the nullspace of higher-priority ones wherever they meet along the graph, not only in action space, ordered by current gradient magnitudes, not a fixed hierarchy. Where two gradients oppose and no weighting of them makes progress, the system instead explores in the nullspace of the stronger one until the conflict dissolves. We robustly solve 100 non-convex navigation and 100 pushT problems, outperforming static potential fields, a diffusion policy, and the same method without projection. Unmodified but embedded in a larger graph, it absorbs perceptual uncertainty, joint limits, and self-collisions on a real robot, solving 49 of 50 pushing trials across varied objects, with active camera control and disturbance recovery emerging from the same coupling. Much of the behavior attributed to planning may thus be within reach of control, once goal representations compose from the conflicts they encounter.
\end{abstract}

\section{Introduction}

\noindent
Robotic behavior is often split between planning and control. Planning is invoked when objectives conflict, since it can anticipate interactions and order subgoals explicitly~\cite{fikes1971strips,garrett2021integrated,toussaint2020describing}, and local reactive control~\cite{brooks1986robust,khatib_unified_1987} is robust but widely held to be prone to local minima in such settings~\cite{gat1998three,koditschek1990robot}.

We argue this misattributes the problem: most local minima~\cite{rimon1992exact} arise not from task complexity but from static encodings that ignore how objectives interact. The designer must define the objectives, but not how to trade them off. That follows from the world's interaction structure---geometry, dynamics, perceptual couplings---which determines whether objectives reinforce or conflict, and changes with the state. Reading it locally, through gradients propagating toward actuation, then suffices to resolve conflicts---as long as conflicts are simultaneous rather than spread over future state sequences, as in Tower of Hanoi or a Rubik's cube~\cite{korf_planning_1987,barret_characterizing_1993}.

Such propagation through a graph-based world model was recently shown to suffice for sequential tasks~\cite{mengers_no_2025}, but simultaneous objectives still conflict in ways myopic gradient following cannot escape. We extend that representation, Active InterCONnect (AICON)~\cite{mengers_no_2025,martin-martin_coupled_2019}, with nullspace projections. Unlike nullspace~\cite{dietrich_overview_2015} or related geometric projections~\cite{ratliff_riemannian_2018,van2022geometric}, which arbitrate engineered priorities in action or configuration space, we resolve conflicts \emph{where} and \emph{when} they arise: lower-priority gradients enter the nullspace of higher-priority ones at whichever intermediate representation they meet, ordered by the interaction structure read out at that moment. Where no weighting resolves the conflict, the nullspace of the stronger gradient instead provides an exploration direction until the structure changes.

\begin{figure}[t]
	\centering
	\definecolor{contact}{RGB}{30,102,245}
	\definecolor{avoid}{RGB}{232,93,12}
	\definecolor{uncert}{RGB}{23,146,153}
	\definecolor{neutral}{RGB}{92,95,119}
	\colorlet{contactavoid}{black!18!contact!50!avoid}
	\colorlet{allthree}{uncert!34!contactavoid}
	\def\pw{0.288\columnwidth}
	\def\gap{0.04\columnwidth}
	\begin{tikzpicture}[
		img/.style={inner sep=0pt, outer sep=0pt},
		toplabel/.style={font=\footnotesize, inner xsep=0pt, inner ysep=1pt,
			align=center, text width=\pw},
		botlabel/.style={font=\scriptsize\linespread{0.88}\selectfont, inner xsep=0pt, inner ysep=1pt,
			align=center, text width=\pw, text=neutral},
		motion/.style={draw=#1, line width=2.5pt, line cap=round,
			-{Stealth[length=6pt,width=8pt]}},
		]

		\node[img] (a) {\includegraphics[width=\pw,viewport=12 454 362 814,clip]{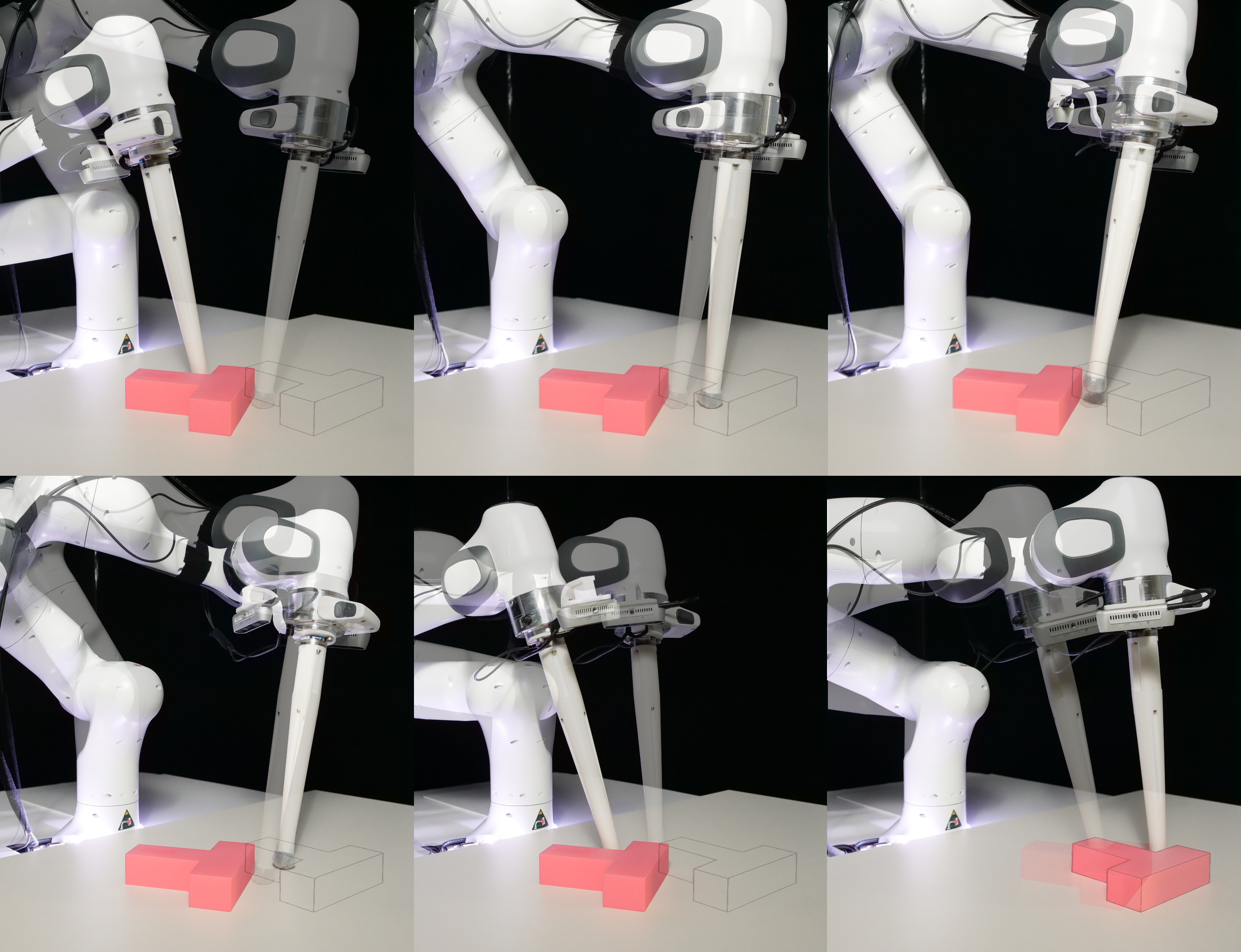}};
		\node[img,right=\gap of a] (b) {\includegraphics[width=\pw,viewport=385 454 735 814,clip]{figures/title_fig_draft.JPG}};
		\node[img,right=\gap of b] (c) {\includegraphics[width=\pw,viewport=760 454 1110 814,clip]{figures/title_fig_draft.JPG}};

		\node[toplabel,text=contact,above=2pt of a] (la) {Seek Good Contact};
		\node[toplabel,text=avoid,  above=2pt of b] (lb) {Avoid Bad Contact};
		\node[toplabel,text=uncert, above=2pt of c,yshift=-1.5pt] (lc) {Reduce Uncertainty};

		\imgcoord{a}{0.658}{0.135}{aFrom}
		\imgcoord{a}{0.471}{0.202}{aTo}
		\imgcoord{b}{0.594}{0.133}{bFrom}
		\imgcoord{b}{0.727}{0.108}{bTo}
		\imgcoord{c}{0.859}{0.723}{cFrom}
		\imgcoord{c}{0.587}{0.840}{cTo}

		\draw[motion=contact] (aFrom) to[bend left=16]  (aTo);
		\draw[motion=avoid]   (bFrom) to[bend right=10] (bTo);
		\draw[motion=uncert]  (cFrom) to[bend left=42]  (cTo);

		\imgcoord{c}{0.620}{0.25}{tqa}
		\imgcoord{c}{0.706}{0.25}{tqb}
		\imgcoord{c}{0.845}{0.92}{tqc}
		\imgcoord{c}{0.703}{0.92}{tqd}
		\begin{scope}
			\clip (tqa) -- (tqb) -- (tqc) -- (tqd) -- cycle;
			\node[img] at (c.center)
			{\includegraphics[width=\pw,viewport=760 454 1110 814,clip]{figures/title_fig_draft.JPG}};
		\end{scope}

		\node[anchor=north east, inner sep=0pt] at ([yshift=-6pt]c.south east) (graph) {%
			\begin{tikzpicture}[
				every node/.style={anchor=center},
				state/.style ={circle,    fill=aicBlue,   minimum size=4.2mm, inner sep=0pt},
				inter/.style ={rectangle, fill=aicGreen,  minimum size=4.2mm, inner sep=0pt,
					rounded corners=1.1pt},
				act/.style   ={circle,    fill=aicOrange, minimum size=5.4mm, inner sep=0pt,
					text=white, font=\scriptsize},
				goal/.style  ={diamond,   fill=aicRed,    minimum size=5.8mm, inner sep=0pt,
					text=white, font=\scriptsize},
				link/.style={draw=aicGray, line width=1.6pt, line cap=round, line join=round},
				grad/.style={draw=#1, line width=2.1pt, line join=round, line cap=butt,
					-{Stealth[length=4.2pt,width=5pt]}},
				shaft/.style={draw=#1, line width=2.1pt, line join=round, line cap=butt,
					shorten >=4pt},
				]
				\coordinate (ca)  at (-0.12, 0.00);
				\coordinate (cK)  at ( 0.89, 0.00);
				\coordinate (cS1) at ( 1.74, 0.33);   \coordinate (cS2) at ( 1.74,-0.33);
				\coordinate (cI1) at ( 2.59, 0.33);   \coordinate (cI2) at ( 2.59,-0.33);
				\coordinate (cT1) at ( 3.43, 0.52);   \coordinate (cT2) at ( 3.43, 0.00);
				\coordinate (cT3) at ( 3.43,-0.46);
				\coordinate (cg)  at ( 4.34, 0.00);

				\draw[link] (ca) -- (cK);
				\draw[link] (cK) -- (cS1);  \draw[link] (cK) -- (cS2);
				\draw[link] (cS1) -- (cI1); \draw[link] (cS2) -- (cI2);
				\draw[link] (cI1) -- (cT1); \draw[link] (cI1) -- (cT2);
				\draw[link] (cI2) -- (cT2); \draw[link] (cI2) -- (cT3);
				\draw[link] (cI1) -- (cT3);
				\draw[link] (cT1) -- (cg);  \draw[link] (cT2) -- (cg); \draw[link] (cT3) -- (cg);

				\node[act]   (A)  at (ca)  {$a$};
				\node[inter] (K)  at (cK)  {};
				\node[state] (S1) at (cS1) {};  \node[state] (S2) at (cS2) {};
				\node[inter] (I1) at (cI1) {};  \node[inter] (I2) at (cI2) {};
				\node[state] (T1) at (cT1) {};  \node[state] (T2) at (cT2) {};
				\node[state] (T3) at (cT3) {};
				\node[goal]  (G)  at (cg)  {$g$};

				\draw[grad=contact] (G) -- (cT1) -- (I1);
				\draw[grad=avoid]   (G) -- (cT2) -- (I1);
				\draw[grad=uncert]  (G) -- (cT3) -- (cI2) -- (cS2) -- (K);

				\draw[grad=contactavoid] (I1) -- (cS1) -- (K);
				\draw[shaft=contact]     (I1) -- (cS1) -- (K);
				\draw[shaft=avoid, dash pattern=on 1.8pt off 1.8pt] (I1) -- (cS1) -- (K);

				\draw[grad=allthree] (K) -- (A);
				\draw[shaft=contact] (K) -- (A);
				\draw[shaft=avoid,  dash pattern=on 1.8pt off 3.6pt] (K) -- (A);
				\draw[shaft=uncert, dash pattern=on 1.8pt off 3.6pt, dash phase=3.6pt] (K) -- (A);
			\end{tikzpicture}%
		};

		\node[anchor=west, text width=0.36\columnwidth, align=center,
		font=\footnotesize, text=neutral, inner sep=0pt,xshift=-5pt,yshift=-2pt]
		at (a.west |- graph) (mecha)
		{Composed Through Adaptive Projection and Exploration in Nullspaces Across the AICON Graph};

		\node[fit=(graph)(mecha), inner sep=0pt, draw=none] (midband) {};

		\node[img, anchor=north west] at ([yshift=-6pt]midband.south -| a.west) (d)
		{\includegraphics[width=\pw,viewport=12 26 362 386,clip]{figures/title_fig_draft.JPG}};
		\node[img, right=\gap of d] (e)
		{\includegraphics[width=\pw,viewport=385 26 735 386,clip]{figures/title_fig_draft.JPG}};
		\node[img, right=\gap of e] (f)
		{\includegraphics[width=\pw,viewport=760 26 1110 386,clip]{figures/title_fig_draft.JPG}};

		\node[botlabel, below=2pt of d] (ld) {%
			\textcolor{avoid}{Avoiding Bad Contact}\\ While\\ \textcolor{uncert}{Reducing Uncertainty}};
		\node[botlabel, below=2pt of e] (le) {%
			Resolving Conflict of\\ \textcolor{avoid}{Avoiding Bad Contact}\\ \textcolor{contact}{Seeking Good Contact}};
		\node[botlabel, below=2pt of f] (lf) {%
			\textcolor{contact}{Seeking Good Contact}\\ While\\ \textcolor{uncert}{Reducing Uncertainty}};

		\draw[-{Stealth[length=6.5pt,width=9pt]},line width=3.4pt,draw=neutral!45]
		([xshift=-8pt]d.east |- e.center) -- ([xshift=8pt]e.west |- e.center);
		\draw[-{Stealth[length=6.5pt,width=9pt]},line width=3.4pt,draw=neutral!45]
		([xshift=-8pt]e.east |- f.center) -- ([xshift=8pt]f.west |- f.center);

		\begin{scope}[on background layer]
			\node[fit=(a)(la), fill=contact!9, rounded corners=2pt, inner sep=2pt] {};
			\node[fit=(b)(lb), fill=avoid!9,   rounded corners=2pt, inner sep=2pt] {};
			\node[fit=(c)(lc), fill=uncert!9,  rounded corners=2pt, inner sep=2pt] {};
			\node[fit=(d)(ld), fill=neutral!8, rounded corners=2pt, inner sep=2pt] {};
			\node[fit=(e)(le), fill=neutral!8, rounded corners=2pt, inner sep=2pt] {};
			\node[fit=(f)(lf), fill=neutral!8, rounded corners=2pt, inner sep=2pt] {};
		\end{scope}

	\end{tikzpicture}
	\vspace*{5pt}
	\caption{Multiple reactive objectives (top, arrows mark the motion each one induces) remain active simultaneously. Their gradients propagate through a single AICON graph (middle) and meet wherever the objectives interact, so that their combination and relative priorities adapt to the current interaction structure (bottom), as demonstrated for object pushing (\Cref{sec:pushing}).}\label{fig:titlefig}
\end{figure}

We demonstrate this in two domains where simultaneous conflicts are central: navigation around non-convex obstacles (\Cref{sec:nav2d}) and planar pushing of non-convex objects (\Cref{sec:pushing}), both usually left to planning~\cite{rimon1992exact,perugini_pushing_2025} or learned policies~\cite{chi2024diffusionpolicy} and largely unsolvable by steepest descent without projections~\cite{mengers_no_2025}. Our unified representation handles perception, action, and conflict resolution in both; and a real robot with further constraints needs only added estimators and goals. Ablations isolate both claims: restricting projection to joint space costs performance, as does fixing the priority order.

These results question whether the boundary between planning and control reflects problem structure or merely the limitations of static goal representations. Once replaced by representations that evolve with the state and resolve conflicts as they arise, the distinction loses much of its necessity, connecting robotic behavior generation to broader principles in optimization and biological control (\Cref{sec:discussion}).

\section{Related Work}

\noindent
Approaches to multi-objective robotic behavior span planning, learned policies, and control. While both planning and learning can coordinate subgoals, they often lack the ability to react to disturbances not anticipated or not reflected in the data. Control is inherently reactive and robust but prone to local minima unless supported by hand-crafted hierarchies, switching logic, or heuristics. We build on control, extending it to resolve conflicts between simultaneous objectives online and avoid minima without planning or fixed hierarchies.

\subsection{Planning}\label{sec:rw:planning}

\noindent
Classical planning~\cite{fikes1971strips}, task and motion planning~\cite{garrett2021integrated,toussaint2020describing}, and trajectory optimization~\cite{posa_direct_2014} explicitly balance multiple objectives, but require accurate models and often become prohibitively slow as task complexity grows. For long-horizon problems they frequently serialize subgoals or offload conflict resolution to the skill level~\cite{garrett2021integrated}, effectively sidestepping the multi-goal challenge. Extensions using large {(vision-) language} models to replace search still rely on sequencing tasks~\cite{driess2023palme}, but require anticipating all relevant interactions in advance. As a result, they remain brittle in dynamic and uncertain environments due to their limited reactivity.

\subsection{Learned Policies}\label{sec:rw:learning}

\noindent
Reinforcement~\cite{kalashnikov2018scalable} and imitation learning~\cite{chi2024diffusionpolicy} instead encode goal trade-offs within a single policy, enabling smooth execution and robustness to error cases represented in the data. However, once trained, these trade-offs are fixed: changing priorities or constraints typically requires costly retraining~\cite{dulac2021challenges}. Moreover, recent analysis suggests learned policies often behave like high-dimensional lookup tables of their training data~\cite{he2025demystifying}, making them brittle to novel goal combinations and environmental changes~\cite{dulac2021challenges,ghosh2021generalization}.

\subsection{Control Methods}\label{sec:rw:control}

\noindent
Reactive control exploits local gradient information~\cite{khatib_unified_1987}, offering robustness and low computational cost, but conflicting goals can create local minima that prevent progress~\cite{rimon1992exact}, mitigated by heuristics such as wall-following only in narrow settings~\cite{van1992wall}. Composing multiple objectives is therefore usually organized by an explicit structure: switching and hierarchy as in subsumption~\cite{brooks1986robust}, state machines, or behavior trees~\cite{iiovino2025comparison}; projection as in nullspace control~\cite{dietrich_overview_2015}, Riemannian Motion Policies~\cite{ratliff_riemannian_2018}, or geometric fabrics~\cite{van2022geometric}; or constrained optimization as in hierarchical quadratic programs~\cite{escande2014hierarchical} and control barrier functions~\cite{ames2019control}. However, their priority structures are fixed by the designer rather than read from the current state, effectively hard-coding a policy.

Recent work~\cite{mengers_no_2025} instead made the potential field adaptive, avoiding minima in sequential tasks, but simultaneous goals still interfere. We extend these adaptive potential fields to the simultaneous setting: by projecting gradients into one another’s nullspaces and dynamically adjusting priorities, we resolve conflicts, escape local minima, and retain the robustness of reactive control---without the anticipatory rigidity of planning or the fixed trade-offs of learned policies.

\section{Adaptive Nullspace Projections in AICON}\label{sec:methods_generic}

\noindent
Active InterCONnect (AICON) is a framework for encoding regularities between sensory inputs, actions, and time. By representing the world as a graph of recursive estimators coupled through state-dependent, differentiable interconnections, it supports robust perception~\cite{martin-martin_coupled_2019}, active sensing, and sequential behavior without planning~\cite{mengers_no_2025} 
within a unified representation. Here, we extend this framework to resolve conflicting simultaneous objectives through nullspace projections, addressing the fundamental limitation of myopic gradient descent when goals are in conflict.

We first summarize how AICON encodes and leverages world regularities for perception and action (\Cref{sec:met_recap}), then introduce nullspace projections (\Cref{sec:met_spaces}), and finally show how these projections dynamically resolve conflicts, preventing (\Cref{sec:met:continuous}) and escaping (\Cref{sec:met:escaping}) local minima while ensuring continuity of the control signal~(\Cref{sec:met:stability}).

\subsection{Encoding Regularities for Perception and Action}\label{sec:met_recap}

\noindent
AICON represents the structure of the world by encoding regularities in estimators for relevant quantities $\mathbf{x}_1, \mathbf{x}_2, \dots, \mathbf{x}_N$ of the world state and \emph{active interconnections} that couple them. Each estimator recursively updates its internal state $\mathbf{x}$ based on sensors, actuators, and constraints from the interconnections. These interconnections are differentiable and state-dependent, enabling bidirectional information flow.

Goals are expressed as differentiable scalar cost functions $g(\mathbf{x})$ defined over a component's estimate $\mathbf{x}$. Gradients of these goals can be propagated backward through the network of interconnected components using the chain rule (\Cref{eq:chain}) along a path $\mathrm{p}: g \leadsto \mathbf{a}$ until they reach actuation signals $\mathbf{a}$.
\begin{equation}
	\nabla^\mathrm{p}_\mathbf{a}\;g=\frac{\partial g(\mathbf{x}_1)}{\partial \mathbf{x}_1}\frac{\partial \mathbf{x}_1}{\partial \mathbf{x}_2} \dots \frac{\partial \mathbf{x}_{N-1}}{\partial \mathbf{x}_N}\frac{\partial \mathbf{x}_{N}}{\partial \mathbf{a}} \label{eq:chain}
\end{equation}

Because interconnections are state-dependent and gradients can propagate along multiple paths through the network, the resulting gradient expansion naturally reveals subgoals: adjustments that change the regime of an interconnection and thereby enable further progress toward the goal. In this way, we obtain a gradient from each path $\mathrm{p} \in \mathbb{P}$ connecting the goal $g$ to an actuation signal $\mathbf{a}$. These gradients represent both direct routes toward the goal and intermediate subgoals that steer the system into more favorable regimes.

To generate sequential behavior as in~\cite{mengers_no_2025}, we simply execute the steepest gradient at every time step (\Cref{eq:steepest}, where $\mathbf{k}$ is a control gain). This ensures robust progress even when some paths stall at local stationary points.
\begin{equation}
	\mathbf{a}_{t+1} = \mathbf{a}_{t} - \mathbf{k} \cdot \nabla^{\mathrm{p}^*}_\mathbf{a}\; g \qquad \mathrm{p}^* = \argmax_{\mathrm{p} \in \mathbb{P}} \|\nabla^\mathrm{p}_\mathbf{a}\; g\| \label{eq:steepest}
\end{equation}

This mechanism allows AICON to jointly perform perception and action selection for sequential, feedback-driven behavior~\cite{mengers_no_2025}. However, it cannot cope with conflicting goals and their gradients. To address this limitation, we leverage the same interaction structure and its gradients but replace the steepest-selection rule (\Cref{eq:steepest}) with a combination of gradients through their nullspaces, as described next.

\subsection{Nullspace Projections for Conflict Resolution}\label{sec:met_spaces}

\noindent
When the gradients along multiple paths are simultaneously relevant, they might interfere, leading to conflicting updates in action space. A common strategy to mitigate this is to project the lower-priority gradient onto the subspace that does not conflict with the higher-priority one---its nullspace. 

Formally, given a gradient $\nabla^{\mathrm{p}_1}_\mathbf{q}\, g$ along path $\mathrm{p}_1$ that relates a quantity $\mathbf{q}$ to a goal's error signal, we define a projection operator $\mathcal{N}_1$ that removes interfering components of the gradient $\nabla^{\mathrm{p}_2}_\mathbf{q}\, g$ along a second path $\mathrm{p}_2$. This yields a conflict-free combined gradient:
\begin{equation}\label{eq:proj}
		\nabla^{\mathrm{p}_{1 \rhd 2}}_\mathbf{q}\, g = \nabla^{\mathrm{p}_1}_\mathbf{q}\, g + \mathcal{N}_1\,  \nabla^{\mathrm{p}_2}_\mathbf{q}\, g\,.
\end{equation}

In the general nullspace formulation, this projector $\mathcal{N}_1$ is computed using the pseudoinverse $\mathbf{J}_1^{\#}$ of the respective Jacobian (\Cref{eq:null}) along $\mathrm{p}_1$. Formulations based on QR decomposition (\Cref{eq:qr}) avoid the inversion. In AICON, as in many optimization settings, all goals are scalar costs, allowing us to simplify the projector further to orthogonalization:
\begin{align}
	\label{eq:null}
	\mathcal{N}_1 &= \mathbf{I} - \mathbf{J}_1^{\#}\mathbf{J}_1\\[-1pt]
	&= \mathbf{I} - \mathbf{Q}_1\mathbf{Q}_1^\mathrm{T} 
	& \text{using}\quad \mathbf{J}_1^\mathrm{T} = \mathbf{Q}_1\mathbf{R}_1	\label{eq:qr}\\[-1pt]
	&= \mathbf{I} -\mathbf{v}_1 \mathbf{v}_1^\mathrm{T} 
	& \text{using}\quad \mathbf{Q}_1 = \mathbf{v}_1 = \nicefrac{\nabla^{\mathrm{p}_1}_\mathbf{q}\, g}{\| \nabla^{\mathrm{p}_1}_\mathbf{q}\, g \|}&\,.\label{eq:ortho}
\end{align}

While the projector itself follows from standard nullspace control~\cite{dietrich_overview_2015}, its application here differs in two key ways: projections occur \emph{where} conflicts arise, at any intermediate representation throughout the graph rather than only in action space; and the priority ordering is not fixed in advance but read continuously from current gradient magnitudes, so it changes \emph{when} the interaction structure does.

\subsection{Resolving Goal Conflicts and Preventing Local Minima}\label{sec:met:continuous}

\begin{figure*}[t]
	\centering
	\includegraphics[width=\textwidth]{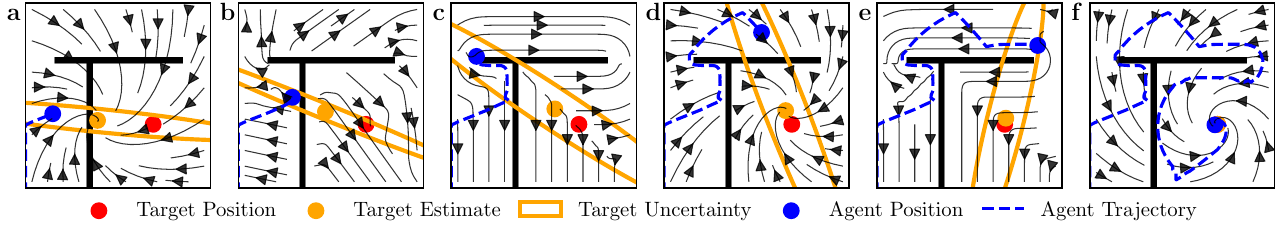}
	\caption{Adaptive conflict resolution solves 2D navigation in a non-convex obstacle configuration that static potential fields cannot solve without prior knowledge of the obstacle. To visualize the evolving potential (arrows), we sample the gradient across the space given the current projections; note that only the local gradient is computed during execution. The composite potential field evolves continuously as gradients are projected into one another's nullspaces. \textbf{(a)} The agent moves toward the target, reducing uncertainty by triangulating in the nullspace of the goal gradient. \textbf{(b)} Attractive and repulsive gradients begin to oppose each other as the agent approaches the obstacle, potentially forming a local minimum. \textbf{(c--e)} Nullspace explorations move the agent along the obstacle boundary until the conflict weakens and a viable path opens. \textbf{(f)} Normal goal pursuit resumes and the agent reaches the target.}\label{fig:2dnav_quali}
	\vspace*{-6pt}
\end{figure*}

\noindent
While projecting one gradient orthogonal to another (\Cref{eq:proj,eq:ortho}) prevents direct interference, AICON leverages this mechanism across multiple interconnected components (\Cref{eq:chain}) to resolve conflicts more broadly. At each state space along the path, gradients can be projected onto one another, yielding a layered resolution process: gradients of progressively lower-priority goals are projected into the orthogonal complements of higher-priority ones wherever they begin to overlap. This continuous resolution of conflicts enables many concurrent goals to be accommodated robustly.

Since gradient magnitudes reflect both sensitivity and the system's ability to
reduce a given cost, they allow us to determine priorities dynamically. At every
quantity $\mathbf{q}$, we assign priorities greedily: the steepest gradient
first, as in \Cref{eq:steepest}, then each subsequent one by its magnitude after
projection into the nullspace of all previously selected gradients:
\begin{gather}\label{eq:ordering}
	\nabla^{\rhd}_{\mathbf{q}} g = \sum_{i=1}^{m}
	\mathcal{N}_{<i}\, \nabla^{\mathrm{p}_{[i]}}_{\mathbf{q}} g\,, \qquad
	m = \min\bigl(\dim(\mathbf{q}), |\mathbb{P}_\mathbf{q}|\bigr)\,,\\[-4.8pt]
	\mathrm{p}_{[i]} = \argmax_{\mathrm{p} \in \mathbb{P}_\mathbf{q} \setminus \{\mathrm{p}_{[j]}\}_{j<i}}
	\bigl\| \mathcal{N}_{<i}\, \nabla^{\mathrm{p}}_{\mathbf{q}} g \bigr\|\,, \qquad
	\mathcal{N}_{<i} = \prod_{k=1}^{i-1} \mathcal{N}_{[k]}\,.
\end{gather}
Here, $\mathbb{P}_\mathbf{q}$ is the set of paths along which goal gradients reach $\mathbf{q}$ and $\mathrm{p}_{[i]}$ the path with the $i$-th highest priority. Each $\mathcal{N}_{[k]}$ follows \Cref{eq:ortho} with $\mathbf{v}_{[k]}$ the normalized \emph{projected} gradient, so the selected directions are orthonormal. Once $\dim(\mathbf{q})$ gradients are selected, the space is exhausted and all remaining candidates are discarded.

Because the summed terms are mutually orthogonal, $\nabla^{\rhd}_\mathbf{q} g$ is free of conflicts. Crucially, this projection happens at the first quantity where the conflicting gradients meet---the representation whose state both goals seek to change. Once resolved there, the combined signal propagates onward as a single gradient, which is projected again only where it meets gradients of further goals. This also makes the exhaustion of $\mathbf{q}$ meaningful: the dimensions consumed are those of the representation the conflicting goals actually contest. Resolving further downstream instead, e.g.\ in joint space, arbitrates between projections of a conflict whose structure is no longer visible. 

Importantly, the notion of a ``goal'' is not restricted to explicitly defined cost functions: every gradient propagating through the network---whether from distinct cost functions or different paths of the same one---is treated uniformly.

\subsection{Escaping Local Minima}\label{sec:met:escaping}

\noindent
Some conflicts admit no resolution by linear combination: when two gradients oppose, every weighting of them stalls, as for a target behind a non-convex obstacle. The classic heuristic here is wall-following~\cite{van1992wall}, which we reinterpret as exploration \emph{in the nullspace of the obstacle gradient}: motion orthogonal to the conflict leaves it untouched but changes the state until the structure producing it no longer holds. 

We generalize this. A conflict is detected wherever two gradients are both strong and near-opposed---cosine similarity below a threshold, neither clearly dominating in magnitude. The system then explores in the nullspace of the dominant gradient, in the direction most consistent with its recent motion, which breaks directional symmetry and biases continuation over reversal. Exploration ends once the gradients are no longer opposed or one dominates. Entry and exit use different thresholds, so the system does not immediately fall back into exploration after leaving it. As gradients propagate through components, this cycle can occur at any intermediate representation, not only in action~space.

Both thresholds can simply be fixed; one setting suffices for all our experiments. Letting them instead follow the system's own progress---the straightness of the recent trajectory of the conflicted quantity, over a short, medium, and long timescale---triggers exploration before gradients are fully opposed and ends it once the system is demonstrably escaping, while preserving the entry-exit asymmetry that prevents jitter. While this introduces timescales instead of thresholds, these are scale-free: unlike thresholds on gradient magnitudes, they do not depend on how cost functions are parametrized and transfer across parameterizations.

\subsection{Ensuring Continuity Across Priority Transitions}\label{sec:met:stability}

\noindent
Since priorities follow from projected gradient magnitudes (\Cref{eq:ordering}), small state changes can flip the ordering and make the
control signal discontinuous. We therefore accept a reordering only once a
competing gradient exceeds the current one by a margin, and low-pass filter the
result. Magnitudes also depend on how cost functions are parametrized,
so inconsistent scaling can yield correct orderings but without appropriate control authority; passing them through a softmax before projection preserves
relative ordering while compressing absolute differences, decoupling
\emph{which} gradient dominates from \emph{how much} control authority that
implies. Choosing cost functions can then focus on qualitative priority relations, and reparametrizing one does not destabilize the others. None of these parameters requires careful tuning. They shape how sharply transitions occur, not what the system does, as we verified by sweeping the margin, softmax temperature, and filter constant across 100 \emph{pushT} trials each, where success remained within 94--100\%.

\section{Conflict Resolution in 2D Navigation}\label{sec:nav2d}

\noindent
Navigation in cluttered environments is a canonical testbed for studying multi-objective conflicts in robotics. Classical potential field methods often fail in these settings: non-convex obstacles induce local minima where the attractive gradient to the goal directly opposes repulsive gradients from obstacles, requiring either precomputed planning~\cite{rimon1992exact,kavraki2002probabilistic} or heuristic strategies such as wall-following~\cite{van1992wall}.

This task isolates adaptive conflict resolution: where obstacles induce opposing gradients, nullspace projection and exploration escape local minima (\Cref{fig:2dnav_quali}).

\subsubsection*{Model}\label{sec:nav2d:model}

We simulate a 2D agent with planar position $\mathbf{x}_\mathrm{rob} \in \mathbb{R}^2$ and velocity control $\mathbf{a} \in \mathbb{R}^2$. The agent receives bearing-only measurements to the target $\mathbf{z}_\mathrm{tar}\in \mathbb{R}^1$ and range measurements to the nearest obstacle $\mathbf{z}_\mathrm{obs}\in \mathbb{R}^1$. AICON encodes three estimates: the agent’s position $\mathbf{x}_\mathrm{rob}$, the distance to the closest obstacle $\mathbf{x}_\mathrm{obs}$, and a Gaussian belief over the target's location $\mathbf{x}_\mathrm{tar} \sim \mathbb{N}(\boldsymbol{\mu}_\mathrm{tar}, \boldsymbol{\Sigma}_\mathrm{tar})$ relative to the agent.

\subsubsection*{Goals and Gradients}\label{sec:nav2d:goal}

The primary objective is minimizing the expected distance to the goal while avoiding collisions with obstacles. We define scalar cost functions on the relative target location and the distance to obstacles:
\begin{equation*}
	g_1 = \|\boldsymbol{\mu}_\mathrm{tar}\| + \sqrt{\Tr(\boldsymbol{\Sigma}_\mathrm{tar})} \geq \mathbb{E}[\|\mathbf{x}_\mathrm{tar}\|] \,,
	\quad 
	g_2 = p_\text{collision}(\mathbf{x}_\mathrm{obs})\,.
\end{equation*}
where $p_\text{collision}$ is the likelihood of collision, decreasing with distance and representing a repulsive potential. Gradients of these costs are propagated through AICON’s interconnections to the actuation signals, leading to three distinct gradient paths that reduce distance to the target ($\mathrm{p}_1$), uncertainty over the target distance ($\mathrm{p}_2$), and likelihood of collision ($\mathrm{p}_3$):
\begin{gather*}
	\nabla_\mathbf{a}^{\mathrm{p}_1} g_1=
	\frac{\partial g_1}{\partial \boldsymbol{\mu}_\mathrm{tar}}
	\frac{\partial \boldsymbol{\mu}_\mathrm{tar}}{\partial \mathbf{x}_\mathrm{rob}} 
	\frac{\partial \mathbf{x}_\mathrm{rob}}{\partial \mathbf{a}}\,,
	\quad 
	\nabla_\mathbf{a}^{\mathrm{p}_2} g_1=
	\frac{\partial g_1}{\partial \boldsymbol{\Sigma}_\mathrm{tar}}
	\frac{\partial \boldsymbol{\Sigma}_\mathrm{tar}}{\partial \mathbf{x}_\mathrm{rob}}
	\frac{\partial \mathbf{x}_\mathrm{rob}}{\partial \mathbf{a}}\,,
	\\[-2pt]
	\nabla_\mathbf{a}^{\mathrm{p}_3} g_2=
	\frac{\partial g_2}{\partial \mathbf{x}_\mathrm{obs}}
	\frac{\partial \mathbf{x}_\mathrm{obs}}{\partial \mathbf{x}_\mathrm{rob}}
	\frac{\partial \mathbf{x}_\mathrm{rob}}{\partial \mathbf{a}}\,.
\end{gather*}

\subsubsection*{Qualitative Behavior}\label{sec:nav2d:qual}

Behavior emerges as a sequence of continuous adjustments rather than preplanned trajectories, as shown in \Cref{fig:2dnav_quali}: the agent triangulates toward the target, encounters an opposing gradient near the obstacle, explores until the conflict resolves, and reaches the goal.

\begin{figure}[b]
	\centering
	\includegraphics[width=\linewidth]{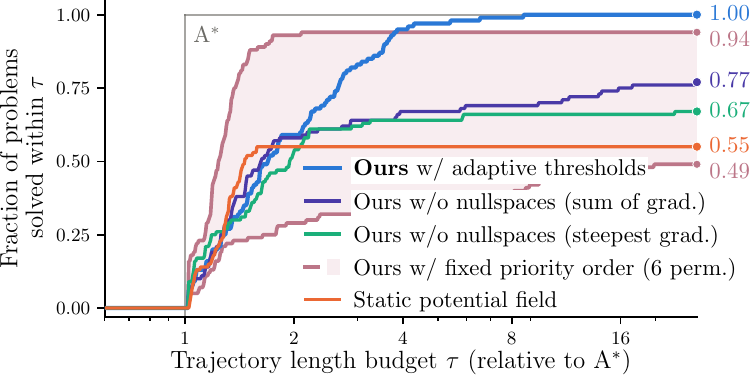}
	\caption{Adaptive priorities solve navigation problems that no fixed alternative does, at the cost of path optimality. Across 100 problems, success is determined by whether a method resolves the conflicts that non-convex obstacles induce. The ablations follow only the steepest gradient or the sum of all gradients, isolating projection and nullspace exploration from the non-static field alone. The band spans all six fixed priority permutations of the three gradients $\nabla_\mathbf{a}^{\mathrm{p}_1} g_1$, $\nabla_\mathbf{a}^{\mathrm{p}_2} g_1$, $\nabla_\mathbf{a}^{\mathrm{p}_3} g_2$: the best is more direct than ours but does not solve every configuration, and the worst falls below static fields.}\label{fig:2dnav_quanti}
\end{figure}

\subsubsection*{Quantitative Results}\label{sec:nav2d:quanti}

We evaluate on 100 problems: 20 obstacle maps with 5 start/goal configurations each. Every baseline uses its best attractive-repulsive trade-off factor, so each is evaluated at its strongest. We solve 
$100\%$, whereas static potential fields solve 
$55\%$, failing exactly where non-convex obstacles induce local minima (\Cref{fig:2dnav_quanti}). Ablating nullspace exploration and projection, so that the agent follows only the steepest gradient ($67\%$) or the sum of all gradients ($77\%$), still beats static fields, since the field remains non-static, but falls short: without projection, conflicting directions are never resolved. Fixing the priority order isolates the \emph{when}: the same projections, but an ordering that no longer changes with the state. A well-chosen order rarely needs exploration and reaches the target far more directly than we do, but still leaves 6\% of configurations unsolved, and a poorly chosen one drops to 49\%---below static fields. No single ordering is right everywhere: which objective should yield depends on where the agent stands relative to the obstacle. We trade path optimality for not needing to know the right order in advance.

Navigation isolates the \emph{when}: all gradients meet in the same 2D action space, so the \emph{where} is trivial. We now turn to a domain where conflicts are generated by the agent's own actions and meet at different~representations.

\section{Conflict Resolution in Object Pushing}\label{sec:pushing}

\noindent
Object pushing tests our thesis directly: contacts to pursue, contacts to avoid, and perceptual or kinematic constraints interact to create local minima that must be resolved online. Non-convex objects sharpen this, since the gradient of a target contact may directly oppose that of a contact to avoid, as in the widely used \textit{pushT} task~\cite{chi2024diffusionpolicy}. Using only an approximate dynamics model, AICON resolves these conflicts from arbitrary start configurations (\Cref{sec:push:T}).

We then embed it in a model reflecting the embodiment and perception constraints of a real setup, adding goals such as reducing perceptual uncertainty and respecting joint limits. The contact-based dynamics model is untouched; continuous conflict resolution absorbs them (\Cref{sec:push:real}). Unlike in navigation, these objectives meet at different representations, so both \emph{where} and \emph{when} matter here.

\subsection{pushT: Object Pushing in Simulation}\label{sec:push:T}

\noindent
In \textit{pushT}, a T-shaped object must be pushed into a goal pose in 2D (\Cref{fig:push_method}, green box); it is a standard benchmark in imitation learning~\cite{chi2024diffusionpolicy}. We use it because its non-convex geometry induces structural local minima: approaching from the wrong side, every immediately available contact moves the object further from the goal, forcing the robot around it.

\subsubsection*{Model} 
\begin{figure}[t]
	\centering
	\newlength{\nodedisty}
	\newlength{\nodedistx}
	\setlength{\nodedisty}{0.9cm}
	\setlength{\nodedistx}{2cm}
	\pgfdeclarelayer{bg}
	\pgfsetlayers{bg,main}
	\begin{tikzpicture}[on grid, 
		node distance=\nodedisty and \nodedistx
		]
		
		\node[aicon/sensor] (apusher) {$\mathbf{a}_\mathrm{{pusher}}$};
		\node[aicon/sensor, below=of apusher] (zpusher) {$\mathbf{z}_\mathrm{{pusher}}$};
		\node[aicon/sensor, below=of zpusher] (zkeypoints) {$\mathbf{z}_\mathrm{{keypoints}}$};

		\node[aicon/state, right=of zpusher] (xpusher) {$\mathbf{x}_\mathrm{{pusher}}$};
		
		\node[aicon/state, right=of xpusher] (xkeypoints) {$\mathbf{x}_\mathrm{{keypoints}}$};
		
		\node[aicon/active, below=of xkeypoints] (pointtracking) {Point Tracking\\ and Projection};
		\node[aicon/active, right=of xkeypoints] (registration) {Point\\ Registration};

		\node[aicon/state, above=of xkeypoints, minimum height=\minimumheightinterconnection] (xcontact) {$\mathbf{x}_\mathrm{{contact}}$\\$p_\mathrm{{contact}}$};
		
		\node[aicon/active, left=of xcontact] (geom) {Geometry-Based\\ Contact Model};
		\node[aicon/active, right=of xcontact] (dyn) {Contact-Based\\ Dynamics Model};
		
		\node[aicon/state, left=of pointtracking] (xee) {$\mathbf{x}_\mathrm{{ee}}$};
		\node[aicon/state, above=of dyn, minimum height=\minimumheightinterconnection,font=\linespread{1}\selectfont] (xobject) {$\mathbf{x}_\mathrm{{obj}}$\\$\boldsymbol{\Sigma}_\mathrm{{obj}}$};
		
		\node[aicon/active, below=of xee] (kinematics) {Robot\\ Kinematics};
		
		\node[aicon/state, below=of kinematics] (xjoints) {$\mathbf{x}_\mathrm{{joints}}$};
		\node[aicon/state, below=of registration] (xheights) {$\mathbf{x}_\mathrm{{heights}}$};
		\node[aicon/state, left=of kinematics] (xbody) {$\mathbf{x}_\mathrm{{body}}$};
		\node[aicon/state, below=of xbody] (xmanip) {$\mathbf{x}_\mathrm{{manip}}$};
		
		\node[aicon/sensor, below=of xheights] (zft) {$\mathbf{z}_\mathrm{{ft}}$};
		\node[aicon/sensor, below=of zft] (zrgbd) {$\mathbf{z}_\mathrm{{rgbd}}$};
		\node[aicon/sensor, right=of kinematics] (zjoints) {$\mathbf{z}_\mathrm{{joints}}$};
		\node[aicon/sensor, right=of xjoints] (ajoints) {$\mathbf{a}_\mathrm{{joints}}$};
		
		\node[aicon/goal, yshift=0.0\nodedisty, xshift=0.03\nodedistx] at (xobject.north west) (g1) {$g_1$};
		\node[aicon/goal,yshift=0.0\nodedisty, xshift=0.03\nodedistx] at (xheights.north west) (g2) {$g_2$};
		\node[aicon/goal,yshift=0.0\nodedisty, xshift=0.03\nodedistx] at (xjoints.north west) (g3) {$g_3$};
		\node[aicon/goal,yshift=0.0\nodedisty, xshift=0.03\nodedistx] at (xbody.north west) (g4) {$g_4$};
		\node[aicon/goal,yshift=0.0\nodedisty, xshift=0.03\nodedistx] at (xmanip.north west) (g4) {$g_5$};
		
		\draw[draw=teal, ultra thick, dotted, rounded corners] ([yshift=0.05cm]xobject.east) -- ([yshift=0.05cm]xobject.west) -- ([yshift=-0.35cm]xobject.west) -- ([yshift=-0.35cm]xobject.east) -- cycle;
		
		\node [rectangle, above=1.1\nodedisty of geom, align=center,text=teal, xshift=0.5\nodedistx,font=\normalsize] (sim) {Disembodied Model\\ for \textit{pushT}};
		
		\node [rectangle, below=0.75\nodedisty of xjoints, align=center,text=orange,xshift=-0.2\nodedistx,font=\normalsize] (rw) {Embodied Real-World Model};

		\begin{pgfonlayer}{bg}
			
			\node [right=1\nodedistx of ajoints,yshift=0cm,xshift=-0.1cm,yshift=0.3cm] {\includegraphics[width=2.6cm]{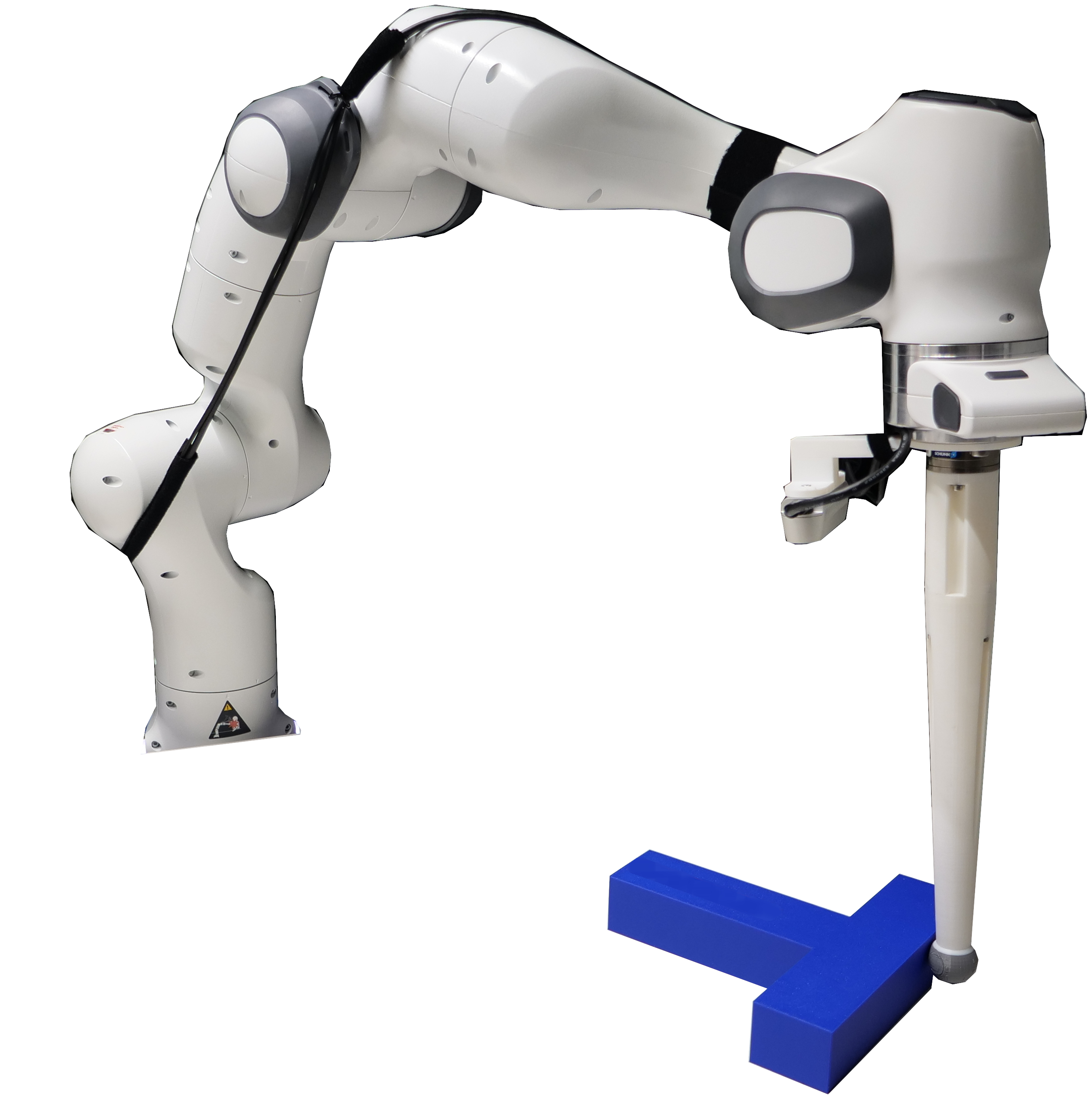}};
			\node [above=\nodedisty of apusher,yshift=-0\nodedisty] {\includegraphics[width=0.7cm]{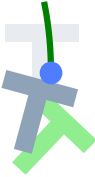}};
			
			\draw[fill=teal, rounded corners, fill opacity=0.01, draw=teal, ultra thick] ([xshift=-0.5\nodedistx,yshift=1.7\nodedisty]apusher.center) -- ([yshift=-0.45\nodedisty, xshift=-0.5\nodedistx]zkeypoints.center) -- ([yshift=-0.45\nodedisty, xshift=0.5\nodedistx]zkeypoints.center) -- ([yshift=0.6\nodedisty, xshift=0.5\nodedistx]zkeypoints.center) -- ([yshift=-0.4\nodedisty, xshift=0.4\nodedistx]xkeypoints.center) -- ([yshift=0.5\nodedisty, xshift=0.4\nodedistx]xkeypoints.center) -- ([yshift=-0.5\nodedisty, xshift=0.6\nodedistx]dyn.center) -- ([yshift=0.7\nodedisty, xshift=0.6\nodedistx]xobject.center) -- cycle;
			
			\draw[fill=orange, rounded corners, fill opacity=0.01, draw=orange, ultra thick] ([xshift=-0.55\nodedistx,yshift=0.5\nodedisty]geom.center) -- ([yshift=-0.5\nodedisty, xshift=0.45\nodedistx]zkeypoints.center) -- ([yshift=-0.5\nodedisty, xshift=-0.5\nodedistx]zkeypoints.center) -- ([yshift=-1.1\nodedisty, xshift=-0.5\nodedistx]xmanip.center) -- ([yshift=-1.1\nodedisty, xshift=0.65\nodedistx]zrgbd.center) -- ([yshift=0.65\nodedisty, xshift=0.65\nodedistx]xobject.center) -- ([yshift=0.65\nodedisty, xshift=-0.5\nodedistx]xobject.center) -- ([yshift=-0.5\nodedisty, xshift=-0.5\nodedistx]xobject.center) -- cycle;
			
			
			\draw[aicon/link] (zpusher.center) -- (xpusher.center);
			\draw[aicon/link] (apusher.center) -- (xpusher.center);
			\draw[aicon/link] (zkeypoints.center) -- ([yshift=0.5\nodedisty,xshift=0.5\nodedistx]zkeypoints.center) --([yshift=-0.5\nodedisty,xshift=-0.5\nodedistx]xkeypoints.center) -- (xkeypoints.center);
			
			\draw[aicon/link] (zrgbd.center) -- (pointtracking.center);
			\draw[aicon/link] (zft.center) -- ([yshift=0.5\nodedisty, xshift=0.55\nodedistx]zft.center) -- ([yshift=-0.6\nodedisty, xshift=0.55\nodedistx]dyn.center) -- (dyn.center);
			\draw[aicon/link] (zft.center) -- (xheights.center);
			\draw[aicon/link] (zjoints.center) -- (kinematics.center);
			\draw[aicon/link] (ajoints.center) -- (kinematics.center);
			
			\draw[aicon/link] (xpusher.center) -- (xee.center);
			\draw[aicon/link] (xpusher.center) -- (geom.center);
			
			\draw[aicon/link] (xcontact.center) -- (geom.center);
			\draw[aicon/link] (xcontact.center) -- (dyn.center);
			\draw[aicon/link] (xobject.center) -- (dyn.center);
			\draw[aicon/link] (xkeypoints.center) -- (geom.center);
			\draw[aicon/link] (xkeypoints.center) -- (registration.center);
			\draw[aicon/link] (xobject.center) -- ([yshift=-0.5\nodedisty, xshift=0.55\nodedistx]xobject.center) --
			([yshift=0.6\nodedisty, xshift=0.55\nodedistx]registration.center) -- (registration.center);
			
			\draw[aicon/link] (pointtracking.center) -- (xkeypoints.center);
			\draw[aicon/link] (pointtracking.center) -- (xee.center);
			\draw[aicon/link] (pointtracking.center) -- (xheights.center);
			
			\draw[aicon/link] (kinematics.center) -- (xee.center);
			\draw[aicon/link] (kinematics.center) -- (xjoints.center);
			\draw[aicon/link] (kinematics.center) -- (xbody.center);
			\draw[aicon/link] (kinematics.center) -- (xmanip.center);
		\end{pgfonlayer}
		
	\end{tikzpicture}
	\caption{By coupling state estimators through active interconnections, we construct a model that solves the pushT task through reactive gradient-based control (green box). When interconnected with additional estimators into a larger model (orange box), the same approach solves object pushing on a real robot. The real-world model introduces additional goals, including perceptual uncertainty reduction and joint limit avoidance, that our approach seamlessly incorporates through the same conflict resolution mechanism.}\label{fig:push_method}
\end{figure}

We model pushT in AICON using a minimal set of observations and actions: the pusher’s 2D location $\mathbf{z}_\mathrm{pusher} \in \mathbb{R}^2$, the object’s $N$ corner keypoints $\mathbf{z}_\mathrm{keypoints} \in \mathbb{R}^{N\times2}$, and the pusher’s velocity as actuation signal $\mathbf{a}_\mathrm{pusher} \in \mathbb{R}^2$. From these, we estimate the pusher position $\mathbf{x}_\mathrm{pusher}$, object keypoints $\mathbf{x}_\mathrm{keypoints}$, and the object pose $\mathbf{x}_\mathrm{obj}$ (centroid and orientation). Possible contacts $\mathbf{x}_\mathrm{contact}\in\mathbb{R}^{N\times2}$ are defined as the closest points on each of the $N$ object sides, with likelihoods $p_\mathrm{contact}$ decreasing with distance (geometry-based contact model in \Cref{fig:push_method}). Object motion under contact is approximated by translation along the contact normal, combined with rotation proportional to the angle between the normal and the vector from contact to centroid. The net effect is a likelihood-weighted combination of all possible contacts (contact-based dynamics model in \Cref{fig:push_method}). Compared to analytic~\cite{perugini_pushing_2025} or data-driven~\cite{bauza_data-efficient_2018} models, ours is deliberately coarse: it captures generic pushing regularities rather than object-specific dynamics, enabling transfer across objects.

\begin{figure*}[t]
	\centering
	\includegraphics[width=\textwidth]{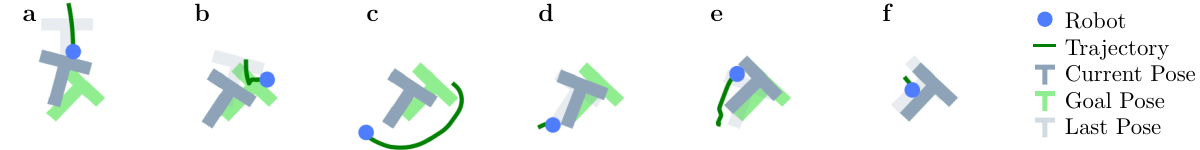}
	\vspace*{-13pt}
	\caption{Representative rollout of AICON on the pushT task. \textbf{(a)} The robot positions itself to increase contact likelihood with the object while modulating contact position in the nullspace. \textbf{(b)} Model imprecision causes the object to be pushed too far without sufficient rotation; the robot reduces contact likelihood to avoid compounding the error. \textbf{(c)} Useful contacts now lie only on the opposite side, creating directly opposing gradients; the system enters nullspace exploration rather than stalling. \textbf{(d)} The robot navigates to the far side of the object. \textbf{(e)} A second conflict arises at a corner contact and is resolved through a brief additional exploration phase. \textbf{(f)} The robot pushes the object into the target pose.}\label{fig:push_quali}\vspace*{-10pt}
\end{figure*}
\begin{figure}[b]
	\centering
	\begin{tikzpicture}
		\node(img) {	\includegraphics[width=\linewidth]{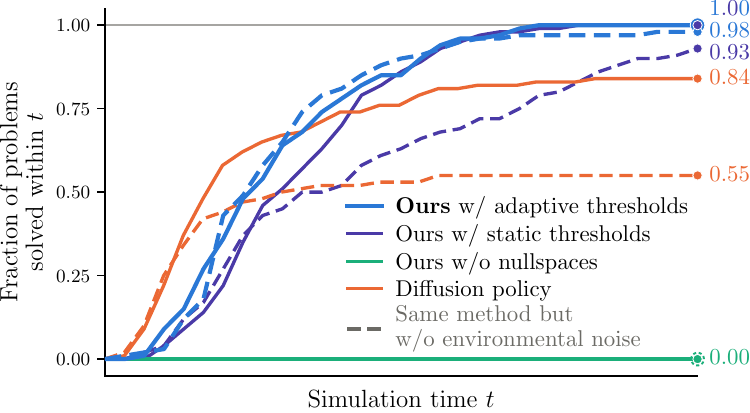}};
		\node[xshift=3.65cm,yshift=-0.7cm] (anno) {\parbox{2cm}{\scriptsize (both sum\\[-3pt] or steepest)}};
	\end{tikzpicture}
	\caption{Conflict resolution, not the specific escape thresholds, is what solves
		pushT: both threshold variants solve all one-hundred randomized configurations
		under noise, while the same method without nullspaces solves none. Curves
		give the fraction solved within a given simulation time; dashed lines repeat
		each method without environment noise. Adaptive thresholds mainly buy speed, and
		the residual noise-free failures of our approach are limit cycles rather than inescapable
		minima. Diffusion policy depends on noise far more heavily, dropping to
		$55\%$.}\label{fig:pusht_quanti}
\end{figure}

\begin{figure*}[t]
	\centering
	\includegraphics[width=\textwidth]{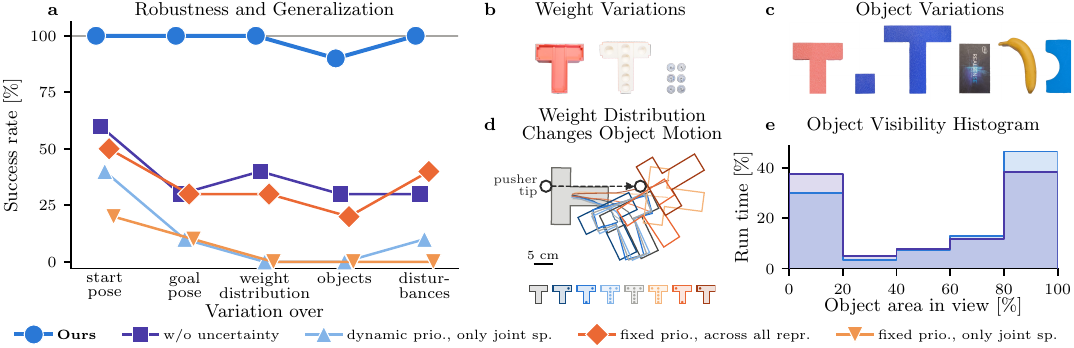}
	\vspace*{-8pt}
\caption{The same representation transfers to a real robot and holds across systematic variation. (\textbf{a}) Success over 50 trials, $N=10$ per variation set: our approach with adaptive thresholds solves 49 of 50. Removing the uncertainty goal costs active camera control and lowers success everywhere; the priority ablations replace adaptive priorities with a fixed order or restrict projection to joint space, showing conflicts must be resolved where they arise. (\textbf{b}) Mass distributions, varied by weight placement inside the object. (\textbf{c}) Object geometries, including non-convex and curved shapes the contact model does not account for. (\textbf{d}) Object trajectories under identical pusher motion for the mass distributions in (\textbf{b}): the spread reflects model error that a planner would have to anticipate but reactive resolution instead absorbs. (\textbf{e}) Fraction of run time at each visibility level. Visibility is not an explicit objective but follows from reducing the pose uncertainty in $g_1$; removing the uncertainty goal lowers visibility and consequently success across all sets (\textbf{a}).}\label{fig:real_quanti}\vspace*{0pt}
\end{figure*}

\subsubsection*{Goal and Gradients}
The task goal is defined as minimizing object-target pose error, $g_1 = \|\mathbf{x}_\mathrm{obj} - \mathbf{x}_\mathrm{target}\|$.
This yields two gradient paths per contact: one through its likelihood ($\mathrm{p}_1$) and one through its position ($\mathrm{p}_2$):
\begin{align*}
	&\nabla_{\mathbf{a}_\mathrm{pusher}}^{\mathrm{p}_1}\, g_1=
	\frac{\partial g_1}{\partial \mathbf{x}_\mathrm{obj}}
	\frac{\partial \mathbf{x}_\mathrm{obj}}{\partial p_\mathrm{contact}}
	\frac{\partial p_\mathrm{contact}}{\partial \mathbf{x}_\mathrm{pusher}}
	\frac{\partial \mathbf{x}_\mathrm{pusher}}{\partial \mathbf{a}_\mathrm{pusher}}
	\,,\\
	&\nabla_{\mathbf{a}_\mathrm{pusher}}^{\mathrm{p}_2}\, g_1 = 
	\frac{\partial g_1}{\partial \mathbf{x}_\mathrm{obj}}
	\frac{\partial \mathbf{x}_\mathrm{obj}}{\partial \mathbf{x}_\mathrm{contact}}
	\frac{\partial \mathbf{x}_\mathrm{contact}}{\partial \mathbf{x}_\mathrm{pusher}}
	\frac{\partial \mathbf{x}_\mathrm{pusher}}{\partial \mathbf{a}_\mathrm{pusher}}\,.
\end{align*}

\subsubsection*{Qualitative Behavior} 
With these gradients, the projection mechanism (\Cref{sec:met:continuous}), and local-minima escape (\Cref{sec:met:escaping}), AICON generates successful behaviors in pushT. A representative rollout is shown in \Cref{fig:push_quali}.

\subsubsection*{Quantitative Results}
Across one-hundred randomly generated start configurations, both threshold
variants solve every configuration under realistic noise
(\Cref{fig:pusht_quanti}), where success is defined as 95\% area coverage, and episodes terminate on a detected limit cycle or after 1000 frames without contact. The steepest-descent baseline
from~\cite{mengers_no_2025}, identical except that it selects the single
strongest gradient without projection or conflict resolution, solves none:
every run stalls at a structural conflict point. A pre-trained pixel-based diffusion
policy of~\cite{chi2024diffusionpolicy} reaches $84\%$, failing on configurations
outside its training distribution, where progress stalls much as in a local
minimum.

Removing environment noise separates two kinds of failure. Our residual failures
($2\%$ adaptive, $7\%$ static thresholds) are limit cycles rather than inescapable minima:
the conflict is resolved repeatedly and any
perturbation breaks the periodicity. Diffusion policy instead drops to $55\%$,
so noise does not merely break symmetry for it but compensates for missing
coverage. Adaptive thresholds also escape conflicts faster, separating
the two curves well before either~plateaus.
\subsection{Real-World Transfer: Handling Additional Goal Conflicts}\label{sec:push:real}

\noindent
We now transfer the model to a real robot with only on-board sensing (\Cref{fig:push_method}, orange box). The model is unchanged apart from unit scaling; idealized sensing and actuation are replaced by an in-hand RGBD camera, a force-torque sensor, and joint-level proprioception and control. These substitutions require additional estimators and interconnections, and introduce new goals that may conflict with the main objective---a strong testbed for AICON’s conflict resolution.

\subsubsection*{Model}
We conduct our experiments on a Panda robot equipped with a custom pusher, an in-hand RGBD camera, and a force-torque sensor. Proprioception $\mathbf{z}_\mathrm{joints}$ and actuation $\mathbf{a}_\mathrm{joints}$ yield joint states $\mathbf{x}_\mathrm{joints}$, from which the kinematics provide the end-effector pose $\mathbf{x}_\mathrm{ee}$ for the push-stick position, the body frames $\mathbf{x}_\mathrm{body}$ for self-collision checks, and current manipulability $\mathbf{x}_\mathrm{manip}$.  
The camera provides segmented object keypoints $\mathbf{z}_\mathrm{keypoints}$. Depth values yield the relative height difference $\mathbf{x}_\mathrm{heights}$ between pusher and object and a pose uncertainty $\boldsymbol{\Sigma}_\mathrm{obj}$ that grows when the object nears the image border. Force-torque signals $\mathbf{z}_\mathrm{ft}$ further inform contact likelihoods and relative height. The full graph, including all projections, runs at \SI{60}{\hertz}; the orthogonalization in \Cref{eq:ortho} is linear in the number of gradients meeting at each quantity.

\subsubsection*{Goals and Gradients}
The task goal remains minimizing object-target error, now extended to penalize perceptual uncertainty: $g_1 = \|\mathbf{x}_\mathrm{obj} - \mathbf{x}_\mathrm{target}\| + \sqrt{\Tr(\boldsymbol{\Sigma}_\mathrm{obj})}$.
This produces gradients not only for pushing but also for adjusting the camera viewpoint. Additional goals maintain contact height ($g_2 = \|\mathbf{x}_\mathrm{heights}\|$), avoid joint limits ($g_3 = p_\mathrm{limit}(\mathbf{x}_\mathrm{joints})$), prevent self-collisions ($g_4 = p_\mathrm{collision}(\mathbf{x}_\mathrm{body})$), and avoid singularities ($g_5 = p_\mathrm{singularity}(\mathbf{x}_\mathrm{manip})$), each propagating through the kinematics to joint space.

\subsubsection*{Qualitative Behavior}
Overall, pushing behavior remains similar to simulation, but perception and robot constraints introduce new conflicts (\Cref{fig:titlefig}). Keypoint-tracking ambiguities and contact-model imprecision occasionally cause errors, e.g., bumping a corner, from which AICON recovers.

The added goals produce distinctive behavior: uncertainty reduction lifts and centers the object in view, counteracted by the height goal and occasionally causing small vertical oscillations, while the camera continuously reorients toward the object. Joint limits, singularities, and self-collisions rarely dominate but redirect the robot into free directions rather than stalling. Nullspace exploration (\Cref{sec:met:escaping}) is triggered mainly by the 2D pushing dynamics, with only rare interference from the other goals in the joint space.

\subsubsection*{Quantitative Results}
We run 50 trials, $N=10$ per variation set (\Cref{fig:real_quanti}): start pose, goal pose, mass distribution, object geometry, and disturbances---disturbance forces on the robot, object displacement, and camera occlusion---applied by hand over the first five interactions. We solve 49 of 50, counting success as error under \SI{2}{\centi\meter} and~\SI{10}{\degree}.

The ablations in Fig.~\ref{fig:real_quanti} \textbf{a} locate this success. Removing the uncertainty goal, and the camera motion it induces, lets the object drift out of view and lowers success. The two priority ablations separate \emph{where} conflicts are resolved from \emph{when} the ordering changes. Restricting projection to joint space costs most: even with adaptive priorities, success falls to 12\% on average, since contact, perceptual, and kinematic conflicts become indistinguishable once projected onto the joints alone. Resolving each conflict at the representation it actually contests lets even a fixed order reach 34\%; reading that order from the current state closes the gap to 98\%, while combining both restrictions performs worst (6\%). Where a conflict is resolved determines whether it can be resolved at all; when the ordering adapts determines whether that resolution survives as the interaction changes.

\Cref{fig:real_quanti} \textbf{d}--\textbf{e} explain why we omit an MPC or planning baseline: object motion under identical pushes diverges with the mass distribution, which the camera cannot observe, so prediction would require a belief over quantities the sensors barely constrain. The same partial observability likely challenges learned policies, whose near-identical observations would demand different actions. We instead resolve the uncertainty online rather than reasoning over it.

Together, these results show that the projection mechanism absorbs additional real-world conflicts without altering the underlying dynamics model, and that resolving each where it arises is what carries the approach to a real robot.

\section{Discussion}\label{sec:discussion}

\subsubsection*{Solvable Problem Classes and Limitations}\label{sec:dis_lim}

Nullspace conflict resolution broadens AICON's solvable class beyond compatible subgoals to concurrent conflicts---given sufficient time. Local, simultaneous conflicts are resolved reactively, but those emerging only through long-term interaction---Tower of Hanoi or a Rubik's Cube, where an early wrong move blocks later progress---remain practically unsolvable. The method ensures eventual progress rather than optimality, and we claim no completeness: exploration is a local escape, so pathological geometries may still trap it.

This mirrors early work on the hierarchy of planning difficulties: conflict-free or nearly serializable tasks are easiest~\cite{barret_characterizing_1993}, while those with mostly invalid sequences or requiring undoing prior subgoals remain hard~\cite{korf_planning_1987, barret_characterizing_1993}. Following Moravec's paradox, though, most robotic manipulation conflicts are likely not of these pathological kinds, but arise between simultaneously relevant goals, where robustness matters more than exact optimality.

\subsubsection*{Connections to Learning and Optimization Approaches}\label{sec:dis_optim}

Conflicting objectives also arise in learning and optimization, trapping progress in local minima~\cite{hu_revisiting_2023}. Multi-objective methods often mitigate this by adjusting updates based on gradient relationships~\cite{yu2020gradient,liu_auto-lambda_2022,zhou_convergence_2022}, since simple linear combinations rarely suffice~\cite{hu_revisiting_2023}. Our results suggest a complementary path: rather than only reshaping update rules, dynamizing the optimization landscape could improve robustness. In our approach, the landscape shifts continuously as priorities change and new gradient paths dominate, making the optimization surface non-stationary by design. This is tractable because the non-stationarity is read from structure rather than imposed: the landscape changes exactly where and when the objectives' interaction does. The constructive role of noise in our pushT results---determinism produces limit cycles that stochasticity resolves---fits the same picture: noise perturbs the effective landscape, breaking periodicity the gradient dynamics cannot escape alone. This parallels noise in SGD~\cite{kleinberg_alternative_2018} but generalizes it: non-stationarity arises not only from injected randomness but from the problem's own structure, suggesting it be treated not as a problem to avoid but as a mechanism to exploit.

\subsubsection*{Connections to Biological Control and Decision-Making}\label{sec:dis_bio}

Nullspace projection also connects to biological control. In human motor control, task-relevant dimensions are stabilized while null dimensions stay free for secondary objectives~\cite{scholz_uncontrolled_1999,latash_motor_2002,rugy_muscle_2012} and are explored during learning~\cite{singh_exploration_2016}; neural population studies show analogous projection-based computations in motor systems~\cite{churchland_neural_2012,flint_long-term_2016} and prefrontal decision-making~\cite{mante_context-dependent_2013,fusi_why_2016, parthasarathy_mixed_2017}. Together with AICON's biological connections~\cite{mengers_no_2025} and its use as a model of biological behavior~\cite{mengers_leveraging_2024,mengers_robotics-inspired_2025,migacev2026planhumanreactiverobotics}, nullspace projection appears not only computationally effective but biologically plausible.

\section{Conclusion}\label{sec:conclusion}

\noindent
We showed that a broad class of multi-objective behaviors typically addressed through planning can instead be generated reactively, by resolving conflicts between simultaneous objectives \emph{where} and \emph{when} they arise: projecting gradients into one another's nullspaces at the representation whose state the conflicting goals contest, ordered by the interaction structure rather than fixed in advance. The extended AICON framework handles tasks that steepest descent over the same representation cannot solve, transfers to a real robot without changing its contact and dynamics model, and accommodates additional constraints through the same mechanism---suggesting that the boundary between planning and control reflects the limitations of static goal representations more than the structure of the tasks themselves. The designer still defines what the robot should achieve; the world determines how those objectives are reconciled.

\bibliographystyle{IEEEtran-nourl}
\bibliography{ICRA26}

\end{document}